\documentclass[11pt,a4paper]{article}
\usepackage[hyperref]{clib_acl}
\setcitestyle{notesep={: }}

\usepackage{times}
\usepackage{latexsym}
\usepackage{booktabs}
\usepackage{amsmath} 
\usepackage{graphics,graphicx}
\usepackage{nicematrix}
\usepackage{tikz}
\usetikzlibrary{fit}

\usepackage[title]{appendix}

\usepackage{microtype}

\usepackage[most]{tcolorbox}

\newtcolorbox{promptbox}[1]{
  colback=gray!5,
  colframe=black!60,
  boxrule=0.4pt,
  arc=1mm,
  left=4pt, right=4pt, top=3pt, bottom=3pt,
  fonttitle=\bfseries\small,
  title=#1,
  fontupper=\small,
  breakable
}

\aclfinalcopy % Uncomment this line for the final camera-ready submission

\title{Reinforcement Learning for improving Large Language Models' Catalan text simplification capabilities}

\author{
Arnau Ayguadé Domingo$^{1,2}$ \qquad
Stefan Bott$^{1}$ \qquad
Horacio Saggion$^{1}$ \\
$^1$Universitat Pompeu Fabra \\
$^2$Barcelona Supercomputing Center \\
\texttt{arnau.ayguade@bsc.es} \qquad
\texttt{\{stefan.bott,horacio.saggion\}@upf.edu}
}

\date{}

\begin{document}
\maketitle
\begin{abstract}
Although automatic text simplification (ATS) is critical for accessibility, its progress has not matched the rapid evolution of broader natural language processing techniques. This paper investigates the application of reinforcement learning (RL) to improve the quality of ATS for low-resource languages using Large Language Models (LLMs). The paper introduces a novel reward function, designed to guide LLMs toward a targeted simplification style with Group Relative Policy Optimization (GRPO), that combines the SARI metric with specific penalty components. The effectiveness of GRPO with this reward function is motivated and demonstrated by post-training IberianLLM-7B-Instruct on the ASSET dataset. After post-training on the English ASSET, the model's ATS performance improves on two curated Catalan benchmarks while also successfully suppressing previously observed negative behaviors. Cross-lingual transfer learning is explored by translating ASSET into Catalan and Spanish and post-training the model on each version, but these fail to show a significant improvement on the out-of-domain benchmark\footnote{Code and final model are available at \url{github.com/arnauad/simplification} and \url{huggingface.co/arnauad/IberianLLM-ASSET-GRPO}}.

\end{abstract}

\section{Introduction}

Automatic text simplification (ATS) is the task of transforming a text into a simpler version, suitable for a target audience, while retaining the original meaning \cite{scarton-specia-2018-learning,saggion2017automatic}. ATS serves a democratizing function by improving access to information in multiple domains, particularly benefiting people with cognitive disabilities, low literacy skills, and second-language learners, thereby reducing communication barriers and promoting greater accessibility \cite{bott-2026}.

Large language models (LLMs) have become the dominant approach for ATS, leveraging their general-purpose instruction-following capabilities to perform simplification without task-specific training \cite{guidroz2025llmbasedtextsimplificationeffect,sanchez-gomez-etal-2025-hulat,zhang-etal-2026-lets}. When the task is narrowly defined, such as sentence simplification, the use of general-purpose LLMs is inefficient, and smaller, task-aligned models are preferred \cite{hayakawa-etal-2025-towards}. Nevertheless, there remains a notable lack of LLMs specifically optimized for ATS. This scarcity is even more pronounced for low-resource languages such as Catalan, which, despite having millions of speakers, suffers from a shortage of sufficiently large datasets and dedicated models. 

Reinforcement learning (RL) is a training technique that adjusts a model's output distribution by increasing the probability of responses that maximize a reward \cite{ziegler2020finetuninglanguagemodelshuman}. Unlike supervised fine-tuning (SFT), which imitates gold-standard simplification pairs, RL optimizes an explicitly chosen objective, allowing the desired properties of a simplification to be encoded directly in a reward function rather than left implicit in the training data. Moreover, RL tends to concentrate probability mass on a narrower set of outputs, resulting in lower output diversity compared to SFT \cite{kirk2024understanding}; we expect this greater consistency to be an advantage for ATS. Despite this potential benefit for readers, RL has received limited attention in ATS \cite{nakamachi-etal-2020-text,yanamoto-etal-2022-controllable}.

Motivated by these limitations, the objective of this work is to improve the performance of LLM-based Catalan sentence simplification and motivate and demonstrate the utility of RL for this task. To do so, we contribute to the field of sentence simplification through the introduction of a novel reward function for Group Relative Policy Optimization (GRPO) \cite{shao2024deepseekmathpushinglimitsmathematical} to guide simplification toward a targeted style. It uses SARI \cite{xu2016optimizing}, as it is language-independent, while incorporating specific constraints to address the limitations of the metric. The effectiveness of GRPO with the proposed reward function for ATS was demonstrated by post-training IberianLLM-7B-Instruct \cite{11401532} on the ASSET dataset \cite{alva-manchego-etal-2020-asset}. The model post-trained on the original English data shows improved SARI alignment with both the ASSET references and the iDEM corpus \cite{bott-2026}, while all post-trained variants suppress the prior over-generation behavior.

Finally, cross-lingual transfer learning was explored by automatically translating ASSET into Spanish and Catalan, and post-training IberianLLM on each translated version. The effect of the training language on Catalan ATS performance is then analyzed.

The paper is organized as follows. Section~\ref{sec:datasets} briefly introduces the two datasets used in this work, outlining the role each one served and the processing steps applied. Section~\ref{sec:methodology} presents the experimental framework and methodological decisions adopted for GRPO-based RL with the novel reward function. Section~\ref{sec:results} presents the results obtained from the experimental evaluation. Section~\ref{sec:conclusions} concludes the paper with the main takeaways of this work, and Section~\ref{sec:limitations} discusses the limitations of the study.

\section{Datasets}
\label{sec:datasets}

This section describes the two main datasets used during this work, iDEM and ASSET, further stating the purpose each one fulfilled and the processing techniques applied to align them with our purpose.

\subsection{iDEM corpus}

The iDEM corpus \cite{bott-2026}, developed under the \textit{Innovative and Inclusive Democratic Spaces for Deliberation and Participation} project, contains expert-curated simplifications in Spanish, Italian, and Catalan. Only the Catalan subset of the corpus was used for this work, consisting of 380 source sentences and their corresponding simplified versions. 

A characteristic of the iDEM corpus is that it includes both document-level simplifications and explanatory transformations, which fall outside the scope of this work, which is sentence simplification. To reduce the presence of sentence pairs whose simplification depended on previous sentences or included explanatory content, BLEURT \cite{sellam-etal-2020-bleurt}, a learned similarity metric, was used to filter the dataset by removing pairs with a BLEURT score below 0.5. As a result, only pairs with a relatively high degree of semantic similarity were retained, increasing the likelihood that the reference simplification could be inferred from the source sentence alone. After filtering, the dataset contained 304 sentence pairs, which were used for benchmarking. 

\subsection{ASSET dataset}

\textit{Abstractive Sentence Simplification Evaluation and Tuning} (ASSET) \cite{alva-manchego-etal-2020-asset} is a crowdsourced, multi-reference English corpus containing sentence-level simplification samples. The dataset comprises 2,000 sentences, each paired with ten human-written simplifications collected through crowdsourcing. The quality of these simplifications is less controlled than those of the expert-curated iDEM corpus; simplifications in ASSET typically involve changing only a few words. In this work, ASSET was chosen for post-training models on the simplification task due to its large number of samples. 

To explore cross-lingual transfer learning, we automatically translated ASSET into Catalan and Spanish using SalamandraTA-7B-Instruct \cite{garcia-gilabert-etal-2025-salamandra}, resulting in three parallel versions of the corpus \cite{conneau2019cross}. 

For each language, only sentence pairs with BLEURT scores between 0.7 and 0.9 were retained. Since ASSET simplifications are already sentence-level, the upper bound was set to exclude near-copy pairs, while the lower bound was applied for consistency with the filtering criterion used on the iDEM corpus. After filtering, the original English dataset retained 15,563 sentence pairs, while the Spanish and Catalan versions retained 13,621 and 12,355, respectively. Although LLMs generally produce high-quality translations, their probabilistic nature may have either amplified or reduced the differences between them. Furthermore, machine translation systems are not designed to preserve the relative complexity of the original texts. Together, these factors may have altered the quality of the samples in the translated corpora.

\section{GRPO Methodology}
\label{sec:methodology}

We employ RL because it allows the optimization objective to be specified explicitly rather than inferred from reference simplifications. In principle, this makes RL well suited to ATS in low-resource settings, where curated parallel corpora are scarce but the properties of a good simplification can be expressed as a scoring function. In practice, this advantage is only fully realized with a reference-free reward; no such metric is currently available for Catalan, so our reward is built on SARI and therefore still depends on reference simplifications. RL methods have also proven particularly effective at mitigating undesirable generation behaviors \cite{shao2024deepseekmathpushinglimitsmathematical,ouyang2022training}, which motivates their use here given an over-generation behavior observed during our exploratory phase.

Group Relative Policy Optimization (GRPO) is a RL algorithm that estimates advantages by normalizing rewards within a group of sampled responses to the same prompt, eliminating the need for a separate value function estimator \cite{shao2024deepseekmathpushinglimitsmathematical}. Outputs with higher relative rewards than the other responses in the same group are made more likely, while those with lower relative rewards are made less likely.

For this work, IberianLLM was post-trained using GRPO on the English, Catalan, and Spanish ASSET datasets independently, resulting in three different versions of the model.

Since the objective of this work was to maximize simplification performance, a brief prompt engineering process was applied prior to the post-training to improve the score of IberianLLM on the filtered iDEM corpus using the SARI metric, as described in Appendix~\ref{app:prompt}. SARI \cite{xu2016optimizing} measures the quality of words that are added, deleted, or kept during simplification compared to golden references.

After GRPO training, all resulting model variants were evaluated on both the Catalan test split of ASSET, representing 10\% of the samples after filtering, and the filtered iDEM corpus. Five independent generations were sampled for each input, and the final performance was computed by averaging the resulting SARI scores.

\subsection{Reward Function}

The reward function is the central component of GRPO, as it directly determines the optimization objective of the policy \cite{zhang-lapata-2017-sentence}. The proposed reward function is based on SARI, normalized to the range $[0,1]$. This metric has important known limitations \cite{alva2021suitability,vasquez-rodriguez-etal-2021-investigating, alfear-etal-2024-meta}, and is adopted out of necessity rather than preference. Reference-free simplification metrics exist for English \cite{huang-kochmar-2024-referee}, but no equivalent is currently available for Catalan. This constrains the setup to a reference-based reward, and consequently to a corpus of reference simplifications, which is precisely the dependency RL would otherwise remove.
In order to minimize the effect of these limitations, two penalty terms are incorporated.

To discourage the model from simply copying the source sentence, the penalty term $P_{\text{copy}}$ is introduced:

\begin{equation}
P_{\text{copy}} =
\begin{cases}
0.5 & \text{if } Y_i = X \ \text{and} \ Y_i \neq Z \\
0 & \text{otherwise}
\end{cases}
\end{equation}

where $X$ is the original sentence, $Z$ the reference simplification, and $Y_i$ the candidate simplification generated by the system.

During the exploratory phase, a tendency toward over-generation was observed for IberianLLM, where it failed to generate its end-of-sequence token and continued producing text until reaching the maximum generation length. Importantly, generations containing over-generated content often achieved higher SARI scores compared with when they were truncated, exposing a significant limitation of the metric. Although the additional text could reduce the DELETE score of SARI, it also increased the likelihood of producing n-grams that matched those added or retained in the reference simplification, thereby improving the ADD and KEEP components. To prevent the LLM from learning to always over-generate, a second penalty term based on the output length is introduced.

Let $L_{\text{pred}}$ denote the word count of the generated output and $L_{\text{ref}}$ the word count of the reference simplification. If the excess length, $\Delta L = L_{\text{pred}} - L_{\text{ref}}$, exceeds a threshold of $\tau = 5$ words, the reward is multiplied by an exponentially decaying penalty controlled by the coefficient $\alpha = 4$. The length penalty term $P_{\text{len}}$ is expressed as follows:

\begin{equation}
P_{\text{len}} =
\begin{cases}
1.0 & \text{if } \Delta L \le \tau \\
\exp\left(-\alpha \cdot \frac{\Delta L}{L_{\text{ref}}}\right) & \text{if } \Delta L > \tau
\end{cases}
\end{equation}

This dynamic penalty scales proportionally with the severity of the length violation relative to the reference length. The coefficient $\alpha$ was chosen to impose a strong penalty, while the threshold $\tau$ allows moderate variation in output length without penalty. 

The final reward function is defined as:

\begin{equation}
    R(X, Y_i, Z) = \max\left(0,\frac{\text{SARI}}{100}-P_{\text{copy}}\right)\cdot P_{\text{len}}
\end{equation}

where $SARI$ and $P_{copy}$ are computed over the source, candidate and reference, while $P_{len}$ depends only on the candidate and reference.

\section{Results}
\label{sec:results}

\begin{figure}
  \centering
  \includegraphics[width=\linewidth]{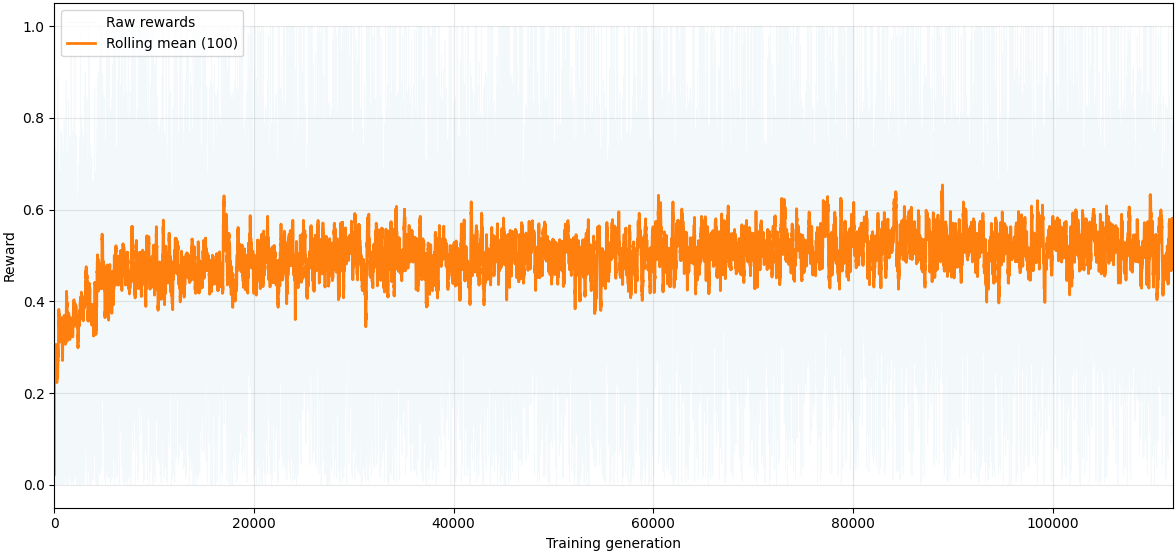}
  \caption{Evolution of the GRPO SARI scores on the original English ASSET dataset. The plot compares the raw SARI scores obtained for each generation with a rolling average computed over a window of 100 steps to highlight the overall optimization trend.}
  \label{fig:trend}
\end{figure}

As shown in Figure \ref{fig:trend}, the SARI scores received by the outputs increased rapidly during approximately the first 10,000 training steps before stabilizing for the remainder of the process. This suggests that the model adapted to the simplification style of the data early in training, and that the number of training samples required may be considerably smaller than the size of the corpora used here. At the same time, it was observed that the percentage of samples triggering the length penalty was drastically reduced, decreasing from almost 50\% to 0-5\% when considering windows of 50 generations.

The same training pattern was observed when post-training on the Catalan and Spanish versions of ASSET, with both runs exhibiting a rapid initial adaptation followed by stable convergence.

The resulting SARI scores on the iDEM benchmark and the Catalan ASSET test set are reported in Table \ref{tab:scores}. Before post-training, the original IberianLLM achieved scores of 43.31 on iDEM and 47.75 on the Catalan ASSET test set.

\begin{table}
\centering
\begin{tabular}{lcccc}
\toprule
\textbf{Benchmark} & \textbf{Original} & \textbf{CA} & \textbf{ES} & \textbf{EN} \\
\midrule
ASSET & 47.75 & \textbf{50.42} & 49.77& \textbf{50.41} \\
iDEM & 43.31 & 42.82 & 42.47& \textbf{44.50} \\
\bottomrule
\end{tabular}
\caption{Catalan text simplification performance of IberianLLM before (Original) and after GRPO post-training on the Catalan (CA), Spanish (ES), and English (EN) versions of ASSET. Average SARI scores are reported for the Catalan ASSET test set and the filtered iDEM benchmark.}
\label{tab:scores}
\end{table}

All post-trained models improved over the original IberianLLM on the Catalan ASSET test set and were no longer suffering from over-generation. All three post-trained models obtained similar results, improving the SARI score by 2–2.7 points. This suggests that all three data versions provided a sufficiently similar learning signal for the model to better match the simplification style represented in ASSET.

On the iDEM benchmark, only post-training on the original English ASSET led to a statistically significant improvement over the original model and also outperformed both the Catalan and Spanish variants.
Neither the Catalan nor the Spanish model differed significantly from the original, and appeared equivalent. Significance was assessed using a bootstrap test on the paired per-item score differences, since all models were evaluated on the same 304 sentences. Full pairwise comparisons and confidence intervals are reported in Appendix~\ref{app:stats}.

\section{Conclusions}
\label{sec:conclusions}

The experiments conducted in this paper indicate that GRPO with the proposed reward function can improve simplification performance, within the limited experimental setting considered here. 
When post-trained on the original ASSET, IberianLLM rapidly learned to suppress over-generation and improved across both evaluation benchmarks. This is notable given that ASSET contains generally milder simplifications than iDEM, so the gain on the more challenging benchmark suggests the model acquired more than the surface style of its training corpus. This did not hold for the models trained on the translated versions, whose in-domain gains did not carry over to iDEM. One possible explanation is that translation altered the simplification characteristics present in the original corpus, though the present results do not isolate this factor.

The extent to which GRPO induces transferable simplification behavior strongly depends on the training data used. This dependency is expected given the setup, since the reward is computed against the reference simplifications of the training corpus, so the properties of those references directly define the optimization target.

Future work should focus on reference-free simplification metrics as reward functions. Such metrics would allow GRPO post-training without manually curated references or unreliable n-gram-based metrics, making the main advantage of RL over other post-training techniques available in practice. Our results point to a possible path even while such metrics remain unavailable for low-resource languages. Post-training on the original English ASSET yielded the only significant improvement on the Catalan iDEM benchmark, which is consistent with simplification ability acquired in English transferring to Catalan, although this rests on a single benchmark and a single training run. An existing English reference-free metric such as REFeREE \cite{huang-kochmar-2024-referee} could therefore be used as a reward on English data, and whether the resulting gains transfer to Catalan and Spanish without curated resources in those languages is a question these experiments motivate rather than answer.

\section{Limitations}
\label{sec:limitations}

The main limitations of this work come from the quality of the ASSET dataset and the SARI metric. 

On the one hand, ASSET was chosen because of its large number of samples. However, since each input is associated with ten reference simplifications, training on this dataset may not reduce the diversity of the model's outputs to the extent originally intended. Similarly, the test split was performed on source-reference pairs rather than on unique source sentences, so a given source sentence may appear in both partitions paired with different references, introducing the risk of data leakage. The results on the iDEM corpus provide a more reliable assessment of simplification performance, and the conclusions regarding cross-lingual transfer rest on that benchmark. Additionally, filtering the datasets by BLEURT score introduces the risk of discarding valid sentence pairs in which the simplification involved substantial content deletion. 

On the other hand, it is known that SARI suffers from many limitations and does not correlate well with human judgments \cite{alva2021suitability,vasquez-rodriguez-etal-2021-investigating}. This metric was chosen due to the lack of reference-free metrics for Catalan, which meant the reward remained tied to reference simplifications and thus constrained the main advantage of RL over SFT.

Finally, due to limited computational resources, each language variant corresponds to a single GRPO training run, and no extensive exploration of different training techniques or hyperparameters was performed and better configurations may exist.

\section{Acknowledgments}

We thank the anonymous reviewers for their insightful comments. We also gratefully acknowledge the \textit{Consorci de Serveis Universitaris de Catalunya} (CSUC) for providing access to the Pirineus III supercomputing infrastructure, which enabled the computational experiments reported in this work. We further acknowledge Universitat Pompeu Fabra for providing the academic environment in which this research was conducted as part of a final degree project. This research is part of the European Union's Horizon Europe research and innovation program under the Grant Agreement No. 101132431 (iDEM: \textit{Innovative and Inclusive Democratic Spaces for Deliberation and Participation}). Views and opinions expressed are, however, those of the authors only and do not necessarily reflect those of the European Union. Neither the European Union nor the granting authority can be held responsible for them.

\bibliographystyle{clib_acl_natbib}
\bibliography{clib}

@inproceedings{nakamachi-etal-2020-text,
    title = "Text Simplification with Reinforcement Learning Using Supervised Rewards on Grammaticality, Meaning Preservation, and Simplicity",
    author = "Nakamachi, Akifumi  and
      Kajiwara, Tomoyuki  and
      Arase, Yuki",
    editor = "Shmueli, Boaz  and
      Huang, Yin Jou",
    booktitle = "Proceedings of the 1st Conference of the Asia-Pacific Chapter of the Association for Computational Linguistics and the 10th International Joint Conference on Natural Language Processing: Student Research Workshop",
    month = dec,
    year = "2020",
    address = "Suzhou, China",
    publisher = "Association for Computational Linguistics",
    url = "https://aclanthology.org/2020.aacl-srw.22/",
    doi = "10.18653/v1/2020.aacl-srw.22",
    pages = "153--159"
}

@inproceedings{yanamoto-etal-2022-controllable,
    title = "Controllable Text Simplification with Deep Reinforcement Learning",
    author = "Yanamoto, Daiki  and
      Ikawa, Tomoki  and
      Kajiwara, Tomoyuki  and
      Ninomiya, Takashi  and
      Uchida, Satoru  and
      Arase, Yuki",
    editor = "He, Yulan  and
      Ji, Heng  and
      Li, Sujian  and
      Liu, Yang  and
      Chang, Chua-Hui",
    booktitle = "Proceedings of the 2nd Conference of the Asia-Pacific Chapter of the Association for Computational Linguistics and the 12th International Joint Conference on Natural Language Processing (Volume 2: Short Papers)",
    month = nov,
    year = "2022",
    address = "Online only",
    publisher = "Association for Computational Linguistics",
    url = "https://aclanthology.org/2022.aacl-short.49/",
    doi = "10.18653/v1/2022.aacl-short.49",
    pages = "398--404"
}

@inproceedings{scarton-specia-2018-learning,
    title = "Learning Simplifications for Specific Target Audiences",
    author = "Scarton, Carolina  and
      Specia, Lucia",
    editor = "Gurevych, Iryna  and
      Miyao, Yusuke",
    booktitle = "Proceedings of the 56th Annual Meeting of the Association for Computational Linguistics (Volume 2: Short Papers)",
    month = jul,
    year = "2018",
    address = "Melbourne, Australia",
    publisher = "Association for Computational Linguistics",
    url = "https://aclanthology.org/P18-2113/",
    doi = "10.18653/v1/P18-2113",
    pages = "712--718"
}

@book{saggion2017automatic,
  author    = {Horacio Saggion},
  title     = {Automatic Text Simplification},
  series    = {Synthesis Lectures on Human Language Technologies},
  volume    = {32},
  year      = {2017},
  publisher = {Morgan \& Claypool Publishers},
  address   = {San Rafael, CA},
  institution = {University of Toronto}
}

@inproceedings{bott-2026,
    title = "A Multilingual Human Annotated Corpus of Original and Easy-to-Read Texts to Support Access to Democratic Participatory Processes",
    author = "Bott, Stefan  and
      Riegler, Verena  and
      Saggion, Horacio  and
      Rascón-Alcaina, Almudena  and
      Khallaf, Nouran",
    editor = "Gurevych, Iryna  and
      Miyao, Yusuke",
    booktitle = "Proceedings of the Language Resources and Evaluation Conference 2026 (LREC 2026)",
    year = "2026",
    address = "Palma, Mallorca, Spain",
    pages = "12",   
}

@misc{guidroz2025llmbasedtextsimplificationeffect,
      title={LLM-based Text Simplification and its Effect on User Comprehension and Cognitive Load}, 
      author={Theo Guidroz and Diego Ardila and Jimmy Li and Adam Mansour and Paul Jhun and Nina Gonzalez and Xiang Ji and Mike Sanchez and Sujay Kakarmath and Mathias MJ Bellaiche and Miguel Ángel Garrido and Faruk Ahmed and Divyansh Choudhary and Jay Hartford and Chenwei Xu and Henry Javier Serrano Echeverria and Yifan Wang and Jeff Shaffer and Eric and Cao and Yossi Matias and Avinatan Hassidim and Dale R Webster and Yun Liu and Sho Fujiwara and Peggy Bui and Quang Duong},
      year={2025},
      eprint={2505.01980},
      archivePrefix={arXiv},
      primaryClass={cs.CL},
      url={https://arxiv.org/abs/2505.01980}, 
}

@inproceedings{sanchez-gomez-etal-2025-hulat,
    title = "{HULAT}-{UC}3{M} at {TSAR} 2025 Shared Task A Prompt-Based Approach using Lightweight Language Models for Readability-Controlled Text Simplification",
    author = "Sanchez-Gomez, Jesus M.  and
      Moreno, Lourdes  and
      Mart{\'i}nez, Paloma  and
      Sanchez-Escudero, Marco Antonio",
    editor = "Shardlow, Matthew  and
      Alva-Manchego, Fernando  and
      North, Kai  and
      Stodden, Regina  and
      Saggion, Horacio  and
      Khallaf, Nouran  and
      Hayakawa, Akio",
    booktitle = "Proceedings of the Fourth Workshop on Text Simplification, Accessibility and Readability (TSAR 2025)",
    month = nov,
    year = "2025",
    address = "Suzhou, China",
    publisher = "Association for Computational Linguistics",
    url = "https://aclanthology.org/2025.tsar-1.15/",
    doi = "10.18653/v1/2025.tsar-1.15",
    pages = "183--192",
    ISBN = "979-8-89176-176-6"
}

@inproceedings{zhang-etal-2026-lets,
    title = "Let{'}s Simplify Step by Step: Guiding {LLM} Towards Multilingual Unsupervised Proficiency-Controlled Sentence Simplification",
    author = "Zhang, Jingshen  and
      Qiu, Xin Ying  and
      Lu, Lifang  and
      Huang, Zhuhua  and
      Hu, Yutao  and
      Wu, Yuechang  and
      Lu, JunYu",
    editor = "Demberg, Vera  and
      Inui, Kentaro  and
      Marquez, Llu{\'i}s",
    booktitle = "Findings of the {A}ssociation for {C}omputational {L}inguistics: {EACL} 2026",
    month = mar,
    year = "2026",
    address = "Rabat, Morocco",
    publisher = "Association for Computational Linguistics",
    url = "https://aclanthology.org/2026.findings-eacl.279/",
    doi = "10.18653/v1/2026.findings-eacl.279",
    pages = "5274--5290",
    ISBN = "979-8-89176-386-9"
}

@inproceedings{hayakawa-etal-2025-towards,
    title = "Towards Trustworthy Lexical Simplification: Exploring Safety and Efficiency with Small {LLM}s",
    author = "Hayakawa, Akio  and
      Bott, Stefan  and
      Saggion, Horacio",
    editor = "Flek, Lucie  and
      Narayan, Shashi  and
      Phương, L{\^e} Hồng  and
      Pei, Jiahuan",
    booktitle = "Proceedings of the 18th International Natural Language Generation Conference",
    month = oct,
    year = "2025",
    address = "Hanoi, Vietnam",
    publisher = "Association for Computational Linguistics",
    url = "https://aclanthology.org/2025.inlg-main.15/",
    pages = "215--231"
}

@misc{shao2024deepseekmathpushinglimitsmathematical,
      title={DeepSeekMath: Pushing the Limits of Mathematical Reasoning in Open Language Models}, 
      author={Zhihong Shao and Peiyi Wang and Qihao Zhu and Runxin Xu and Junxiao Song and Xiao Bi and Haowei Zhang and Mingchuan Zhang and Y. K. Li and Y. Wu and Daya Guo},
      year={2024},
      eprint={2402.03300},
      archivePrefix={arXiv},
      primaryClass={cs.CL},
      url={https://arxiv.org/abs/2402.03300}, 
}

@article{xu2016optimizing,
  title={Optimizing statistical machine translation for text simplification},
  author={Xu, Wei and Napoles, Courtney and Pavlick, Ellie and Chen, Quanze and Callison-Burch, Chris},
  journal={Transactions of the Association for Computational Linguistics},
  volume={4},
  pages={401--415},
  year={2016}
}

@inproceedings{sellam-etal-2020-bleurt,
    title = "{BLEURT}: Learning Robust Metrics for Text Generation",
    author = "Sellam, Thibault  and
      Das, Dipanjan  and
      Parikh, Ankur",
    editor = "Jurafsky, Dan  and
      Chai, Joyce  and
      Schluter, Natalie  and
      Tetreault, Joel",
    booktitle = "Proceedings of the 58th Annual Meeting of the Association for Computational Linguistics",
    month = jul,
    year = "2020",
    address = "Online",
    publisher = "Association for Computational Linguistics",
    url = "https://aclanthology.org/2020.acl-main.704/",
    doi = "10.18653/v1/2020.acl-main.704",
    pages = "7881--7892"
}

@INPROCEEDINGS{11401532,
  author={Lacunza, Iñaki and Saiz, José Javier and Shvets, Alexander and Gonzalez-Agirre, Aitor and Villegas, Marta},
  booktitle={2025 IEEE International Conference on Big Data (BigData)}, 
  title={XDoGE: Multilingual Data Reweighting to Enhance Language Inclusivity in LLMs}, 
  year={2025},
  volume={},
  number={},
  pages={5462-5471},
  doi={10.1109/BigData66926.2025.11401532}}

@inproceedings{alva-manchego-etal-2020-asset,
    title = "{ASSET}: {A} Dataset for Tuning and Evaluation of Sentence Simplification Models with Multiple Rewriting Transformations",
    author = "Alva-Manchego, Fernando  and
      Martin, Louis  and
      Bordes, Antoine  and
      Scarton, Carolina  and
      Sagot, Beno{\^i}t  and
      Specia, Lucia",
    editor = "Jurafsky, Dan  and
      Chai, Joyce  and
      Schluter, Natalie  and
      Tetreault, Joel",
    booktitle = "Proceedings of the 58th Annual Meeting of the Association for Computational Linguistics",
    month = jul,
    year = "2020",
    address = "Online",
    publisher = "Association for Computational Linguistics",
    url = "https://aclanthology.org/2020.acl-main.424/",
    doi = "10.18653/v1/2020.acl-main.424",
    pages = "4668--4679"
}

@inproceedings{garcia-gilabert-etal-2025-salamandra,
    title = "From {SALAMANDRA} to {SALAMANDRATA}: {BSC} Submission for {WMT}25 General Machine Translation Shared Task",
    author = "Garcia Gilabert, Javier  and
      Liao, Xixian  and
      Da Dalt, Severino  and
      Bohman, Ella  and
      Mash, Audrey  and
      De Luca Fornaciari, Francesca  and
      Baucells, Irene  and
      Llop, Joan  and
      Claramunt, Miguel  and
      Escolano, Carlos  and
      Melero, Maite",
    editor = "Haddow, Barry  and
      Kocmi, Tom  and
      Koehn, Philipp  and
      Monz, Christof",
    booktitle = "Proceedings of the Tenth Conference on Machine Translation",
    month = nov,
    year = "2025",
    address = "Suzhou, China",
    publisher = "Association for Computational Linguistics",
    url = "https://aclanthology.org/2025.wmt-1.37/",
    doi = "10.18653/v1/2025.wmt-1.37",
    pages = "614--637",
    ISBN = "979-8-89176-341-8"
}

@article{alva2021suitability,
  title={The (un) suitability of automatic evaluation metrics for text simplification},
  author={Alva-Manchego, Fernando and Scarton, Carolina and Specia, Lucia},
  journal={Computational Linguistics},
  volume={47},
  number={4},
  pages={861--889},
  year={2021},
  publisher={MIT Press One Rogers Street, Cambridge, MA 02142-1209, USA journals-info~…}
}

@article{ouyang2022training,
  title={Training language models to follow instructions with human feedback},
  author={Ouyang, Long and Wu, Jeffrey and Jiang, Xu and Almeida, Diogo and Wainwright, Carroll and Mishkin, Pamela and Zhang, Chong and Agarwal, Sandhini and Slama, Katarina and Ray, Alex and others},
  journal={Advances in neural information processing systems},
  volume={35},
  pages={27730--27744},
  year={2022}
}

@inproceedings{zhang-lapata-2017-sentence,
    title = "Sentence Simplification with Deep Reinforcement Learning",
    author = "Zhang, Xingxing  and
      Lapata, Mirella",
    editor = "Palmer, Martha  and
      Hwa, Rebecca  and
      Riedel, Sebastian",
    booktitle = "Proceedings of the 2017 Conference on Empirical Methods in Natural Language Processing",
    month = sep,
    year = "2017",
    address = "Copenhagen, Denmark",
    publisher = "Association for Computational Linguistics",
    url = "https://aclanthology.org/D17-1062/",
    doi = "10.18653/v1/D17-1062",
    pages = "584--594"
}

@inproceedings{kirk2024understanding,
  title={Understanding the effects of rlhf on llm generalisation and diversity},
  author={Kirk, Robert and Mediratta, Ishita and Nalmpantis, Christoforos and Luketina, Jelena and Hambro, Eric and Grefenstette, Edward and Raileanu, Roberta},
  booktitle={International Conference on Learning Representations},
  volume={2024},
  pages={20620--20653},
  year={2024}
}

@article{conneau2019cross,
  title={Cross-lingual language model pretraining},
  author={Conneau, Alexis and Lample, Guillaume},
  journal={Advances in neural information processing systems},
  volume={32},
  year={2019}
}

@inproceedings{vasquez-rodriguez-etal-2021-investigating,
    title = "Investigating Text Simplification Evaluation",
    author = "V{\'a}squez-Rodr{\'i}guez, Laura  and
      Shardlow, Matthew  and
      Przyby{\l}a, Piotr  and
      Ananiadou, Sophia",
    editor = "Zong, Chengqing  and
      Xia, Fei  and
      Li, Wenjie  and
      Navigli, Roberto",
    booktitle = "Findings of the Association for Computational Linguistics: ACL-IJCNLP 2021",
    month = aug,
    year = "2021",
    address = "Online",
    publisher = "Association for Computational Linguistics",
    url = "https://aclanthology.org/2021.findings-acl.77/",
    doi = "10.18653/v1/2021.findings-acl.77",
    pages = "876--882"
}

@inproceedings{huang-kochmar-2024-referee,
    title = "{REF}e{REE}: A {RE}ference-{FREE} Model-Based Metric for Text Simplification",
    author = "Huang, Yichen  and
      Kochmar, Ekaterina",
    editor = "Calzolari, Nicoletta  and
      Kan, Min-Yen  and
      Hoste, Veronique  and
      Lenci, Alessandro  and
      Sakti, Sakriani  and
      Xue, Nianwen",
    booktitle = "Proceedings of the 2024 Joint International Conference on Computational Linguistics, Language Resources and Evaluation (LREC-COLING 2024)",
    month = may,
    year = "2024",
    address = "Torino, Italia",
    publisher = "ELRA and ICCL",
    url = "https://aclanthology.org/2024.lrec-main.1200/",
    pages = "13740--13753"
}

@inproceedings{alfear-etal-2024-meta,
    title = "Meta-Evaluation of Sentence Simplification Metrics",
    author = "Alfear, Noof Abdullah  and
      Kazakov, Dimitar  and
      Al-Khalifa, Hend",
    editor = "Calzolari, Nicoletta  and
      Kan, Min-Yen  and
      Hoste, Veronique  and
      Lenci, Alessandro  and
      Sakti, Sakriani  and
      Xue, Nianwen",
    booktitle = "Proceedings of the 2024 Joint International Conference on Computational Linguistics, Language Resources and Evaluation (LREC-COLING 2024)",
    month = may,
    year = "2024",
    address = "Torino, Italia",
    publisher = "ELRA and ICCL",
    url = "https://aclanthology.org/2024.lrec-main.981/",
    pages = "11229--11235"
}

@misc{ziegler2020finetuninglanguagemodelshuman,
      title={Fine-Tuning Language Models from Human Preferences}, 
      author={Daniel M. Ziegler and Nisan Stiennon and Jeffrey Wu and Tom B. Brown and Alec Radford and Dario Amodei and Paul Christiano and Geoffrey Irving},
      year={2020},
      eprint={1909.08593},
      archivePrefix={arXiv},
      primaryClass={cs.CL},
      url={https://arxiv.org/abs/1909.08593}, 
}

\begin{appendices}

\section{Prompt Engineering}
\label{app:prompt}

To improve simplification performance prior to RL post-training, four prompting strategies were explored and evaluated using the SARI metric between IberianLLM's generations and the iDEM simplifications. For each configuration, four simplifications were sampled per input to obtain a more reliable score estimate.

\begin{enumerate}
    \item A minimal prompt instructing the model to simplify the input sentence, without additional context or constraints.
    \item The same zero-shot instruction, divided into system and user roles.
    \item Three high-quality examples, manually selected from the iDEM corpus, presented as prior user–assistant turns via the chat template.
    \item Introduction of simplification rules derived from the European Easy-to-Read guidelines. Multiple combinations were tested in the system prompt, combined with the few-shot examples from (3); the combination achieving the highest SARI score was retained.
\end{enumerate}

The rule-based configuration (4) achieved the highest SARI score and was used for post-training.

\begin{promptbox}{System Prompt}
Ets un assistent expert en simplificació de textos en català.
Rebràs una frase en català i hauràs de retornar la versió simplificada en català.

Objectiu: Simplificar frases mantenint EXACTAMENT el mateix significat.

Segueix aquestes pautes:
\begin{itemize}
\setlength\itemsep{0.1em}
    \item No utilitzis paraules difícils. Si has d'utilitzar paraules difícils, assegura't d'explicar-les sempre de manera clara.
    \item No utilitzis metàfores.
    \item Utilitza llenguatge actiu en lloc de passiu.
    \item Repeteix informació important si cal.
\end{itemize}
\end{promptbox}

\begin{promptbox}{Few-shot Examples}
\textbf{Input:} Frase català: El 2004, Barcelona va rebre 4,4 milions de turistes. \\
\textbf{Output:} Simplificació català: El 2004, Barcelona va acollir 4,4 milions de visitants.

\vspace{4pt}
\textbf{Input:} Frase català: Important: aquest visat no és aplicable per a les persones amb ciutadania europea. \\
\textbf{Output:} Simplificació català: És important tenir-ho present perquè aquest visat no s'aplica als ciutadans europeus.

\vspace{4pt}
\textbf{Input:} Frase català: – Edificis d'habitatges (de propietaris únics, de propietaris que destinen els immobles a lloguer, comunitats de veïns, etc.). \\
\textbf{Output:} Simplificació català: – Edificis d'habitatges, com per exemple: cases d'una sola persona propietària, cases que es lloguen, comunitats de veïns i veïnes, i altres.
\end{promptbox}

\begin{promptbox}{User Template}
Simplifica la següent frase en català mantenint el significat i la mateixa llengua.

Respon NOMÉS amb la frase simplificada.

Frase català: \texttt{\{sentence\}} \\
Simplificació català:
\end{promptbox}

\section{iDEM Statistical Significance}
\label{app:stats}

Table~\ref{tab:pairwise} reports the pairwise differences in mean SARI score on the iDEM benchmark for all six comparisons among the four model variants. For each comparison, the difference and 95\% confidence interval were computed with a percentile bootstrap (10,000 resamples) over the per-item score differences, pairing each of the 304 iDEM sentences across the two models being compared. A comparison is marked as significant when its confidence interval excludes zero, corresponding to $p < 0.05$.

\begin{table}[h]
\centering
\small
\begin{tabular}{lcc}
\toprule
\textbf{Comparison} & \textbf{Diff.} & \textbf{95\% CI} \\
\midrule
CA vs. Original & $-0.49$ & $[-1.83, 0.79]$ \\
ES vs. Original & $-0.84$ & $[-2.17, 0.40]$ \\
EN vs. Original* & $+1.19$ & $[0.18, 2.20]$ \\
ES vs. CA        & $-0.34$ & $[-1.06, 0.38]$ \\
EN vs. CA*        & $+1.68$ & $[0.55, 2.89]$ \\
EN vs. ES*        & $+2.02$ & $[1.02, 3.14]$ \\
\bottomrule
\end{tabular}
\caption{Pairwise differences in mean SARI score on the iDEM benchmark, with 95\% paired bootstrap confidence intervals. Asterisks indicate that the comparison is significant. }
\label{tab:pairwise}
\end{table}

Only comparisons involving the English-trained model reach significance. While the confidence interval for the English and Original comparison excludes zero, its lower bound indicates that the true improvement could be quite small, so the result should be read as evidence of a real but possibly modest effect.

The Spanish versus Catalan interval is the narrowest in the table and excludes differences larger than roughly one SARI point in either direction. Within the resolution of this benchmark the two translated variants therefore appear equivalent, and the separation visible in these results is between the translated corpora and the untranslated one rather than between languages. These experiments do not establish why.

\end{appendices}

\end{document}